\newif\ifanonymousversion
\anonymousversionfalse

\documentclass[sigconf]{acmart}

\setcopyright{acmcopyright}
\renewcommand\footnotetextcopyrightpermission[1]{} 

\usepackage{graphicx}
\usepackage{xcolor}
\usepackage{soul}
\usepackage{caption}
\usepackage{subcaption}
\usepackage{multirow}
\usepackage{multicol}
\usepackage{adjustbox}
\usepackage{hyperref}

\newcommand{\nsf}[1]{\href{https://www.nsf.gov/awardsearch/showAward?AWD_ID=#1}{#1}}

\begin{document}

\title[Fall Leaf Adversarial Attack on Traffic Sign Classification]{Fall Leaf Adversarial Attack on Traffic Sign Classification\vspace{1cm}}

\ifanonymousversion
\author{Anonymous Submission}

\else

\author{Anthony Etim}
\affiliation{%
  \institution{Yale University}
  \city{New Haven} 
  \state{CT} 
  \country{USA}
}
\email{anthony.etim@yale.edu}

\author{Jakub Szefer}
\affiliation{%
  \institution{Northwestern University}
  \city{Evanston} 
  \state{IL} 
  \country{USA}
}
\email{jakub.szefer@northwestern.edu}

\fi

\begin{abstract}
    Adversarial input image perturbation attacks have emerged as a significant threat to machine learning algorithms, particularly in image classification setting. These attacks involve subtle perturbations to input images that cause neural networks to misclassify the input images, even though the images remain easily recognizable to humans. One critical area where adversarial attacks have been demonstrated is in automotive systems where traffic sign classification and recognition is critical, and where misclassified images can cause autonomous systems to take wrong actions. This work presents a new class of adversarial attacks. Unlike existing work that has focused on adversarial perturbations that leverage human-made artifacts to cause the perturbations, such as adding stickers, paint, or shining flashlights at traffic signs, this work leverages nature-made artifacts: tree leaves. By leveraging nature-made artifacts, the new class of attacks has plausible deniability: a fall leaf stuck to a street sign could come from a near-by tree, rather than be placed there by an malicious human attacker. To evaluate the new class of the adversarial input image perturbation attacks, this work analyses how fall leaves can cause misclassification in street signs. The work evaluates various leaves from different species of trees, and considers various parameters such as size, color due to tree leaf type, and rotation. The work demonstrates high success rate for misclassification. The work also explores the correlation between successful attacks and how they affect the edge detection, which is critical in many image classification algorithms.
\end{abstract}

\maketitle

\pagestyle{plain}

\section{Introduction}
\label{sec_introduction}

Machine learning algorithms such as Deep Neural Networks (DNNs) are inherently vulnerable to adversarial input image perturbation attacks, where small input modifications can cause a trained model to incorrectly classify the input images~\cite{chakraborty2018adversarial}. For example, Szegedy et al. first discovered that well-performing deep neural networks are susceptible to adversarial attacks~\cite{szegedy2013intriguing}. As these models are increasingly deployed in critical systems such as automated vehicles and self-driving cars, the potential for adversarial manipulation presents a serious threat to the reliability and safety of the automotive vehicle, especially when traffic sign classification and recognition are the targets of the attacks.

Successful adversarial attacks have been demonstrated on both traffic sign classification~\cite{eykholt2018robust} and recognition~\cite{wei2022adversarial} models. In existing work, an attacker can manipulate the entire surface of a traffic sign~\cite{lu2017adversarial}, create  stickers~\cite{eykholt2018robust}, use shadows~\cite{zhong2022shadows} or use light from flashlights~\cite{hsiao2024natural} to disrupt the model's performance. These techniques cause image classification to fail -- however, they all rely on human-made artifacts, such as the stickers, to create an attack. These attacks do not have plausible deniability as, when discovered, it is clear that the change was human-made. Our new attack class, on the other hand, leverages nature-made objects, namely leaves, to create the adversarial perturbations. A fall leaf stuck to a street sign could come from a near-by tree, rather than be placed there by an attacker, thus making it harder to attribute misclassification to be caused by malicious attacks -- and not by random leaf falling on a street~sign.

To evaluate the new class of the adversarial input image perturbation attacks, this work analyses how fall leaves can cause misclassification in street signs. The work considers established LISA street sign data set, and analyzes how what looks like fallen leaves from trees can cause misclassification of the street sign images. The work evaluates various leaves from different species of trees, which at the same time are of different colors. Through extensive evaluation we also consider various parameters such as size and rotation of the leaves. As detailed in Section~\ref{attack_results}, the work demonstrates high success rate for misclassification, with misclassified images having confidence values of 97\%.

To aid in understanding the attacks, this work also explores the correlation between successful attacks and how they affect the edge detection. Different metrics, such as edge length, orientation, intensity, or center of gravity are explored. Their values for successful and unsuccessful attacks are compared and discussed.

\subsection{Contributions}

The contributions of this work are as follows:

\begin{enumerate}

    \item Demonstration of a new class of adversarial input image perturbation attacks based on perturbations due to nature-made objects that are placed on street signs.
    
    \item Evaluation of various attack parameters: leaf size, leaf color due to tree species, and leaf rotation.
    
    \item Analysis of edge detection and how successful attacks impact edge detection vs. unsuccessful attacks.
    
\end{enumerate}

\section{Background}
\label{background}

This section provides an overview of adversarial input perturbation attacks, the LISA dataset of U.S. street sign images, and the LISA-CNN model, used here as a target classifier. Our approach applies LISA-CNN to traffic signs for the adversarial attack evaluation.

\subsection{Adversarial Attacks}

In machine learning, small pixel modifications in input images can significantly alter the DNN predictions, a technique known as adversarial attacks. These attacks operate in two main modes: white-box, where attackers have full access to the model's parameters, and black-box, where attackers lack model knowledge~\cite{liang2022adversarial,chakraborty2018adversarial}.

The Fast Gradient Sign Method (FGSM) \cite{goodfellow2014explaining} is a foundational white-box attack that generates adversarial examples by adjusting the input in the direction of the gradient’s sign, scaled by a small constant. This simple yet effective technique highlights deep learning vulnerabilities by causing misclassifications with minimal input changes, and it has inspired more advanced iterative methods like the Projected Gradient Descent (PGD) attack \cite{mkadry2017towards}.

Adversarial attacks are generally unique to each input image, yet a more scalable approach involves creating a single perturbation pattern applicable across an entire dataset~\cite{moosavi2017universal}. Such universal patterns increase the practical feasibility of adversarial attacks, especially with techniques like adversarial patches, which confine perturbations to specific image areas, further enhancing their applicability in real-world scenarios~\cite{brown2017adversarial}.

To improve patch robustness under spatial shifts, varying camera angles, and environmental factors, methods like Expectation over Transformations (EoT) apply transformations such as rotation, scaling, and brightness adjustments to boost patch resilience during training~\cite{athalye2018synthesizing}. However, Eykholt et al.~\cite{eykholt2018robust} identified limitations in synthetic transformations, proposing the Robust Physical Perturbations (RP2) method, which incorporates real-world variations (e.g., angles, distances, lighting) for more practical deployment.

Emerging techniques, including light and shadow-based attacks, leverage natural phenomena as stealthy adversarial methods. In~\cite{zhong2022shadows}, shadows disrupt image interpretation in digital and physical environments, posing threats by inducing model misclassifications. Similarly, ~\cite{hsiao2024natural} demonstrated the vulnerability of vision systems to natural light sources (e.g., sunlight, flashlights) which, by altering lighting conditions, subtly mislead models, complicating defense strategies in real-world applications.

Our approach leverages natural leaves as adversarial elements, exploiting the realistic appearance and plausible deniability they offer. Unlike typical adversarial perturbations, a leaf on a traffic sign appears as an incidental occurrence, such as a leaf from a nearby tree sticking to the sign. This work demonstrates how such naturally inspired attacks can subtly manipulate image classification, challenging the model’s accuracy by introducing features that seamlessly blend into the environment. By exploring parameters like leaf type, size, color, and rotation, we reveal the effectiveness of these subtle, nature-based adversarial attacks in deceiving machine learning models.

\subsection{LISA-CNN and Traffic Sign Dataset}

The LISA dataset provides a diverse collection of U.S. traffic sign images, encompassing 47 different road sign types~\cite{lisa}. To address dataset imbalance, our work, following previous research, concentrates on the 16 most common signs~\cite{hsiao2024natural}. This selection allows for focused training of convolutional neural networks (CNNs) tailored for traffic sign classification. LISA-CNN, a model with three convolutional layers followed by a fully connected layer, is specifically trained on this subset of the LISA dataset~\cite{eykholt2018robust}. By incorporating images taken under varied environmental conditions and perspectives, the architecture effectively supports the development of models that can generalize well in real-world scenarios. Studies have shown that models trained on the LISA dataset achieve high accuracy and real-time performance in traffic sign recognition, enhancing the capabilities of autonomous vehicle perception systems~\cite{pavlitska2023adversarial}.

In this work, we explore the vulnerability of LISA-CNN to adversarial attacks using natural artifacts like leaves. By strategically placing leaves on signs, we examine how variations in leaf type, size, color, and rotation can mislead the model into misclassifying traffic signs. This approach leverages the realistic and inconspicuous nature of leaves, presenting a unique adversarial attack that blends seamlessly into the environment and highlights the challenges of ensuring robustness in autonomous perception systems.

\begin{figure*}[t]
\begin{subfigure}[b]{0.19\textwidth}
        \centering
        \includegraphics[width=2.2cm]{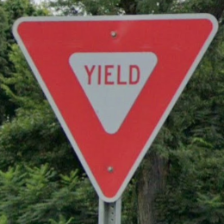}
        \caption{\footnotesize \centering Yield\newline Test Image}
    \label{fig:yield_image}
    \end{subfigure}
    \hfill
    \begin{subfigure}[b]{0.19\textwidth}
        \centering
\includegraphics[width=2.2cm]{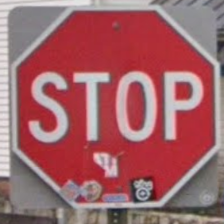}
        \caption{\footnotesize \centering Stop\newline Test Image}
        \label{fig:stop_image}
    \end{subfigure}
    \hfill
     \begin{subfigure}[b]{0.19\textwidth}
        \centering
        \includegraphics[width=2.2cm]{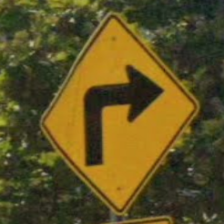}
        \caption{\footnotesize \centering Turn Right\newline Test Image}
    \label{fig:right_image}
    \end{subfigure}
    \hfill
     \begin{subfigure}[b]{0.19\textwidth}
        \centering
    \includegraphics[width=2.2cm]{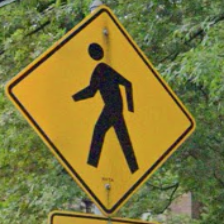}
        \caption{\footnotesize \centering Ped. Crossing\newline Test Image}
    \label{fig:pedestrian_image}
    \end{subfigure}
    \hfill
    \begin{subfigure}[b]{0.19\textwidth}
        \centering
        \includegraphics[width=2.2cm]{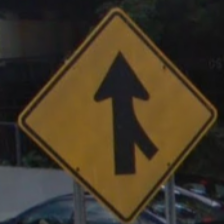}
        \caption{\footnotesize \centering Merge\newline Test Image}
    \label{fig:merge_image}
    \end{subfigure}
    \hfill
    \caption{Test images used in evaluation of the attack.}
    \label{fig:Test_Images}
\end{figure*}

\begin{figure*}[t]
 \begin{subfigure}[b]{0.19\textwidth}
        \centering
        \includegraphics[width=2.2cm]{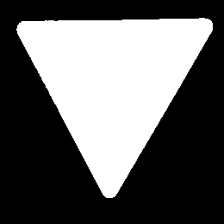}
        \caption{\footnotesize \centering Yield\newline Output Mask}
    \label{fig:yield_output_mask}
    \end{subfigure}
    \hfill
    \begin{subfigure}[b]{0.19\textwidth}
        \centering
\includegraphics[width=2.2cm]{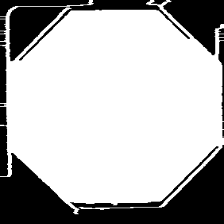}
        \caption{\footnotesize \centering Stop\newline Output Mask}
        \label{fig:stop_output_mask}
    \end{subfigure}
    \hfill
    \begin{subfigure}[b]{0.19\textwidth}
        \centering
        \includegraphics[width=2.2cm]{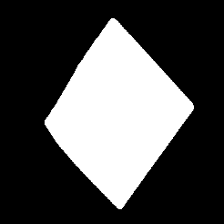}
        \caption{\footnotesize \centering Turn Right\newline Output Mask}
    \label{fig:right_output_mask}
    \end{subfigure}
    \hfill
     \begin{subfigure}[b]{0.19\textwidth}
        \centering
    \includegraphics[width=2.2cm]{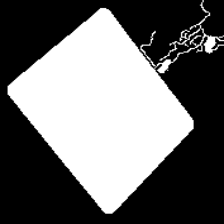}
        \caption{\footnotesize \centering Ped. Crossing\newline Output Mask}
    \label{fig:pedestrian_output_mask}
    \end{subfigure}
    \hfill
    \begin{subfigure}[b]{0.19\textwidth}
        \centering
        \includegraphics[width=2.2cm]{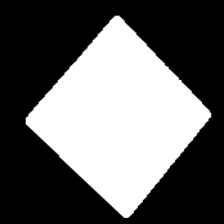}
        \caption{\footnotesize \centering Merge\newline Output Mask}
    \label{fig:merge_output_mask}
    \end{subfigure}
    \hfill
    \caption{Leaf masks generated by our attack method, one for each test image.}
     \label{fig:Output Masks}
\end{figure*}

\section{Threat Model}

In our threat model, we assume that the attacker has partial but limited knowledge of the machine learning system and no access to its internal architecture, weights, or training data, making this a black-box attack scenario. The attacker’s objective is to disrupt the system's performance by exploiting vulnerabilities in the physical environment, specifically through the use of natural artifacts like leaves, which appear innocuous but can mislead the model when strategically positioned.

The attacker can place or adjust leaves on traffic signs, leveraging variations in leaf size, color, type, and orientation to introduce subtle perturbations that are nearly imperceptible to human observers yet sufficient to induce misclassification in the model. The adversary operates in real-world settings such as autonomous vehicles or traffic surveillance systems, where accurate traffic sign recognition is crucial for decision-making. This approach bypasses conventional defenses against adversarial noise by relying on environmental elements that blend seamlessly into natural surroundings, making it both practical and highly evasive. The attacker, however, has no means to influence other external information sources that might aid in traffic sign recognition, nor do they have access to additional digital resources that could directly interact with the model.

\section{Fall Leaf Adversarial Attacks}

This section introduces our novel class of adversarial attacks targeting traffic sign classification systems. Building upon existing techniques, we utilize natural artifacts like leaves as adversarial elements to induce misclassification in traffic sign recognition. The attacks are evaluated on real-life traffic sign images, demonstrating their effectiveness and adaptability in real-world conditions. By examining different leaf types, sizes, colors, and orientations, this study highlights the potential of natural environmental artifacts to disrupt machine learning models used in traffic sign~classification.

\subsection{Leaf Attack}
\label{leaf_attack}

In our approach, leaves are strategically positioned over traffic signs to introduce subtle occlusions that alter the classifier’s perception. By selecting leaf types with varying textures, shapes, and color profiles, we simulate realistic, benign obstructions that nonetheless introduce adversarial perturbations. The attacks are tested on the LISA-CNN  model, a convolutional neural network architecture trained specifically for traffic sign classification tasks. By leveraging the natural diversity in leaves, we create adaptive attacks that exploit model vulnerabilities and assess the LISA-CNN's robustness against occlusions. Experiments were conducted using multiple leaf types and varying their placement to evaluate the model’s potential for misclassification across different types of traffic signs. To the best of our knowledge, this is the first work using leaves to create an adversarial attack on traffic sign recognition.

\subsubsection{Optimal Position Search Using Grid Search}

To systematically identify the best attack location on the traffic sign, we employ a grid search approach~\cite{liashchynskyi2019grid} across the binary mask of the sign. This mask serves as a guide to determine valid regions where the leaf can be placed without exceeding the sign’s boundaries. We divide the sign area into a grid of candidate positions, examining each location to see if the leaf patch fits entirely within the sign mask. For each grid point, the leaf patch is applied, and the modified image is evaluated by the model to check if misclassification occurs.

The grid search is conducted for each combination of patch ratio and rotation angle. At each grid point, the overlayed leaf patch is applied, and the modified image is evaluated for model prediction confidence and classification outcome. If the leaf overlay causes a misclassification, we track the confidence level of the incorrect prediction, identifying the combination of position, patch ratio, and angle that yields the highest confidence in misclassification as the optimal configuration.

\subsubsection{Patch Ratios}

To determine the most effective size of the leaf occlusion, we introduce the concept of patch area ratios. This parameter defines the proportion of the traffic sign's area that the leaf occlusion should cover, ranging from smaller ratios (e.g., 0.1) to larger ones (up to 0.5). Given a traffic sign mask, we calculate the patch area based on each specified ratio and generate a corresponding square patch size. This approach ensures that each leaf occlusion remains proportional to the sign, allowing for a controlled assessment of how occlusion size impacts the LISA-CNN model’s classification performance.

\subsubsection{Rotation Angles}

Considering that leaf positioning in real environments is rarely uniform, we vary the rotation angle of each leaf overlay, creating diverse orientations to better mimic realistic occlusions. For each patch ratio, we test multiple rotation angles (0°, 45°, 90°, 135°, 180°, 225°, 270°, and 315°). By rotating the leaf overlay, we aim to uncover whether specific orientations contribute more effectively to misclassification, providing insight into the model’s sensitivity to occlusions with varied directional patterns.

\subsection{Leaf Images}

For our experiments, we selected three types of leaves commonly found in natural street environments: Maple, Oak and Poplar.

\begin{figure}[t]
\begin{subfigure}[b]{0.16\textwidth}
        \centering
        \includegraphics[width=2.2cm]{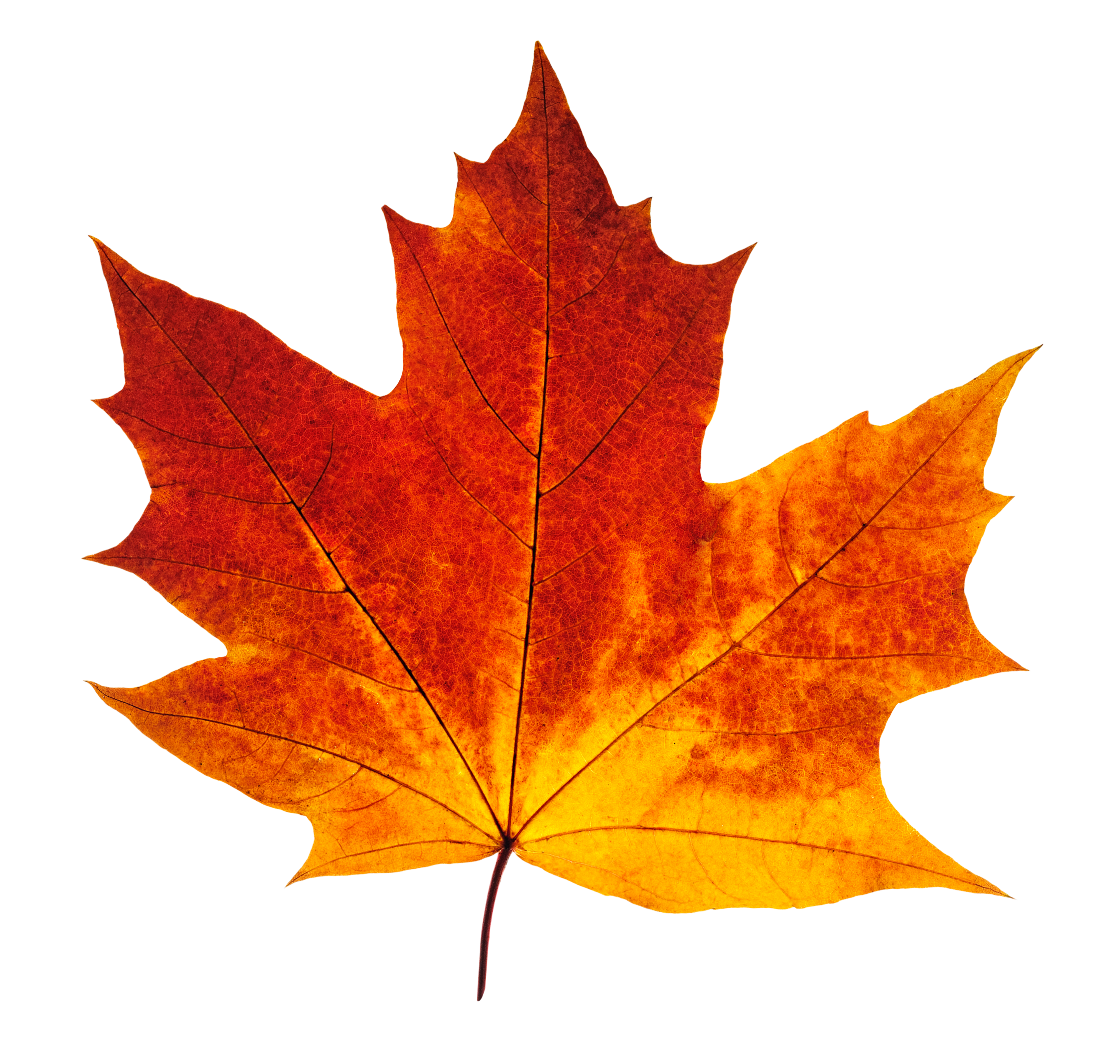}
        \caption{\footnotesize \centering Maple\newline Leaf Image}
    \label{fig:maple_leaf_image}
    \end{subfigure}
    \hfill
    \begin{subfigure}[b]{0.15\textwidth}
        \centering
\includegraphics[width=2.2cm]{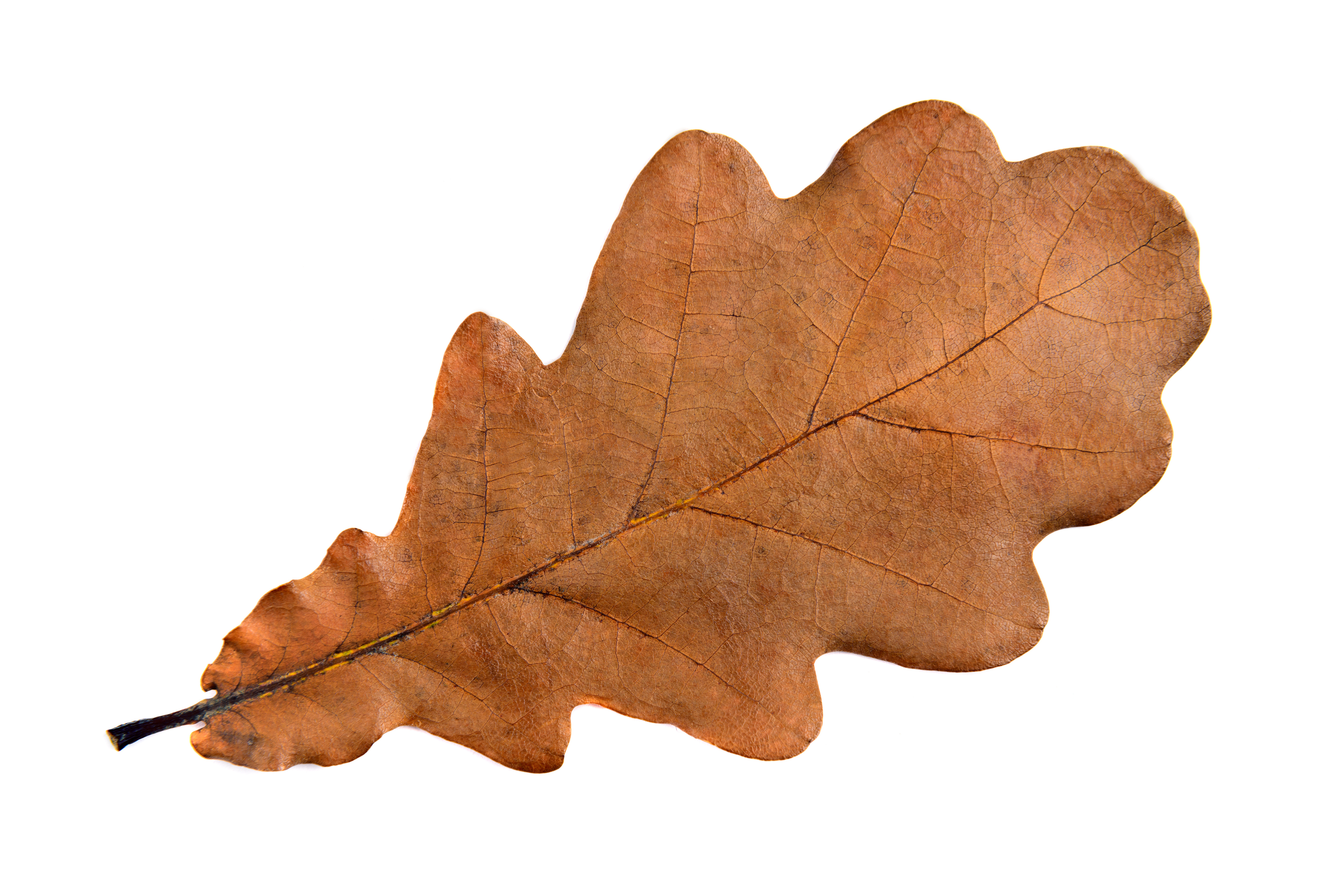}
        \caption{\footnotesize \centering Oak\newline Leaf Image}
        \label{fig:oak_leaf_image}
    \end{subfigure}
    \hfill
     \begin{subfigure}[b]{0.15\textwidth}
        \centering
        \includegraphics[width=2.2cm]{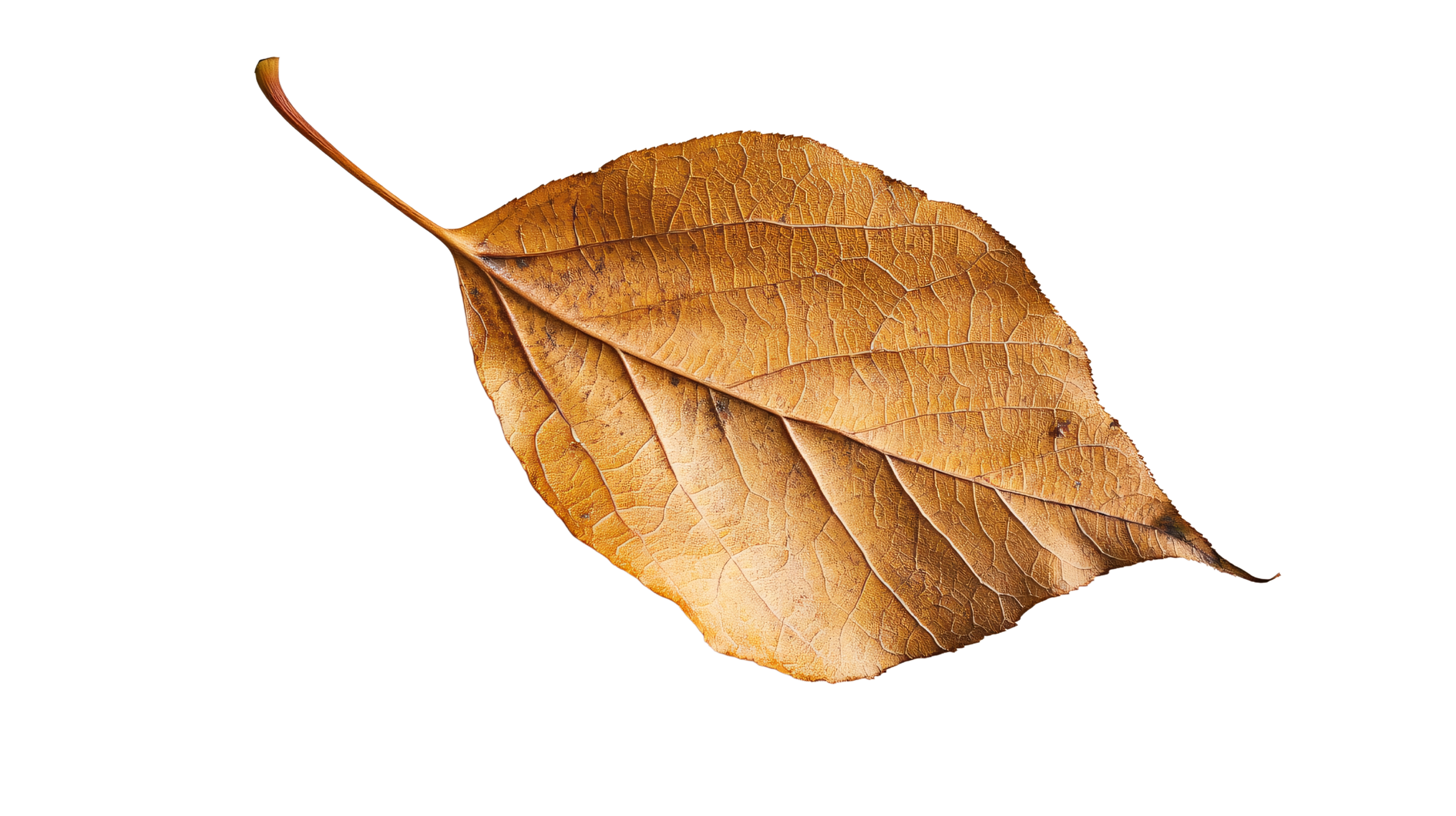}
        \caption{\footnotesize \centering Poplar\newline Leaf Image}
    \label{fig:poplar_leaf_image}
    \end{subfigure}
    \hfill
    \caption{Leaf images used in evaluation of the attacks.}
    \label{fig:Leaf_Images}
\end{figure}

Maple is recognizable for its broad, lobed structure, providing varied coverage and natural texture. Oak features a rounded, lobed shape that can obscure different parts of the sign surface, causing varying levels of misclassification. Polar is known for its smooth, oval-shaped leaves, adding a high-contrast occlusion. These leaves were chosen based on their prevalence and distinct shapes, which allowed us to study the effect of different occlusion patterns and leaf orientations on the LISA-CNN model. The leaf images used can be seen in Figure~\ref{fig:Leaf_Images}.

\subsection{Custom Leaf Masks}

To enhance the natural look and effectiveness of the leaf-based adversarial attack, we developed custom binary masks for each leaf type to precisely control their placement and interaction with the target traffic sign. Unlike general occlusion-based approaches, our method requires detailed masks that align with the edges and contours of each leaf to maintain the appearance of a natural obstruction.

The mask generation process begins by processing the input leaf image, first converting it to grayscale to simplify subsequent steps. To reduce noise, a Gaussian blur~\cite{hummel1987deblurring} is applied, followed by the Canny edge detection algorithm~\cite{rong2014improved}, which highlights the edges within the image. Next, we apply dilation and morphological closing techniques~\cite{gil2002efficient} to close any gaps in the contours, creating a more continuous outline. The contours are then filtered based on area, with the largest contour selected to represent the leaf shape. Finally, this contour is drawn onto a black background to produce a binary mask that accurately captures the leaf's form, allowing it to be effectively overlaid onto traffic sign images for occlusion.

This method enabled us to customize masks for each traffic sign image, regardless of the angle under which the leaf would appear, ensuring that the leaf occlusion was applied in a precise and controlled manner. The flexibility of this approach allows for realistic placement of leaf shapes, enhancing the natural appearance of the various adversarial examples.

\begin{figure*}[t]
 \begin{subfigure}[b]{0.19\textwidth}
        \centering
        \includegraphics[width=2.2cm]{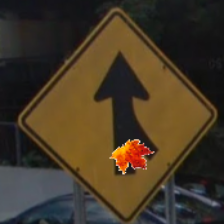}
        \caption{\footnotesize \centering Merge Maple\newline Attack Image}
    \label{fig:merge_maple_attack_image}
    \end{subfigure}
    \hfill
\begin{subfigure}[b]{0.19\textwidth}
        \centering
        \includegraphics[width=2.2cm]{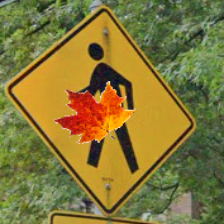}
        \caption{\footnotesize \centering Ped. Crossing Maple\newline Attack Image}
    \label{fig:ped_crossing_maple_attack_image}
    \end{subfigure}
    \hfill
 \begin{subfigure}[b]{0.19\textwidth}
        \centering
\includegraphics[width=2.2cm]{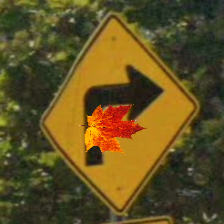}
        \caption{\footnotesize \centering Turn Right Maple\newline Attack Image}
        \label{fig:right_maple_attack_image}
    \end{subfigure}
    \hfill
\begin{subfigure}[b]{0.19\textwidth}
        \centering
        \includegraphics[width=2.2cm]{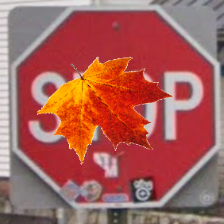}
        \caption{\footnotesize \centering Stop Maple\newline Attack Image}
    \label{fig:stop_maple_attack_image}
    \end{subfigure}
    \hfill
\begin{subfigure}[b]{0.19\textwidth}
        \centering
        \includegraphics[width=2.2cm]{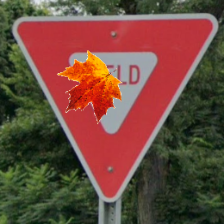}
        \caption{\footnotesize \centering Yield Maple\newline Attack Image}
    \label{fig:yield_maple_attack_image}
    \end{subfigure}
    \hfill
\begin{subfigure}[b]{0.19\textwidth}
        \centering
        \includegraphics[width=2.2cm]{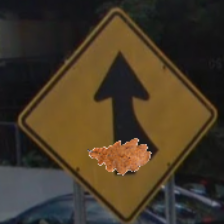}
        \caption{\footnotesize \centering Merge Oak\newline Attack Image}
    \label{fig:merge_oak_attack_image}
    \end{subfigure}
    \hfill
 \begin{subfigure}[b]{0.19\textwidth}
        \centering
        \includegraphics[width=2.2cm]{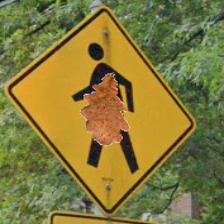}
        \caption{\footnotesize \centering Ped. Crossing Oak\newline Attack Image}
    \label{fig:ped_crossing_oak_attack_image}
    \end{subfigure}
    \hfill
 \begin{subfigure}[b]{0.19\textwidth}
        \centering
        \includegraphics[width=2.2cm]{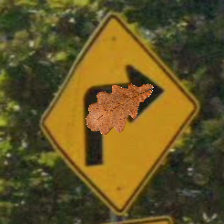}
        \caption{\footnotesize \centering Turn Right Oak\newline Attack Image}
    \label{fig:right_oak_attack_image}
    \end{subfigure}
    \hfill
\begin{subfigure}[b]{0.19\textwidth}
        \centering
        \includegraphics[width=2.2cm]{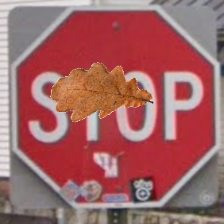}
        \caption{\footnotesize \centering Stop Oak\newline Attack Image}
    \label{fig:stop_oak_attack_image}
    \end{subfigure}
    \hfill
\begin{subfigure}[b]{0.19\textwidth}
        \centering
        \includegraphics[width=2.2cm]{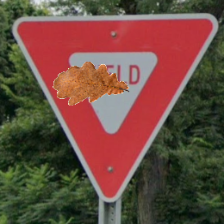}
        \caption{\footnotesize \centering Yield Oak\newline Attack Image}
    \label{fig:yield_oak_attack_image}
    \end{subfigure}
    \hfill
\begin{subfigure}[b]{0.19\textwidth}
        \centering
        \includegraphics[width=2.2cm]{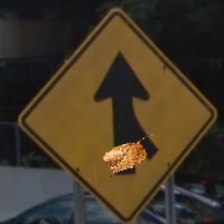}
        \caption{\footnotesize \centering Merge Poplar\newline Attack Image}
    \label{fig:merge_poplar_attack_image}
    \end{subfigure}
    \hfill
\begin{subfigure}[b]{0.19\textwidth}
        \centering
        \includegraphics[width=2.2cm]{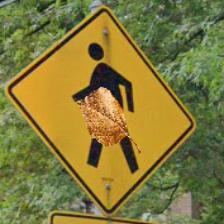}
        \caption{\footnotesize \centering Ped. Crossing Poplar\newline Attack Image}
\label{fig:ped_crossing_poplar_attack_image}
    \end{subfigure}
    \hfill
\begin{subfigure}[b]{0.19\textwidth}
        \centering
    \includegraphics[width=2.2cm]{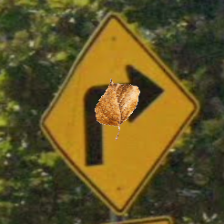}
        \caption{\footnotesize \centering Turn Right Poplar\newline Attack Image}
    \label{fig:right_poplar_attack_image}
    \end{subfigure}
\begin{subfigure}[b]{0.19\textwidth}
        \centering
        \includegraphics[width=2.2cm]{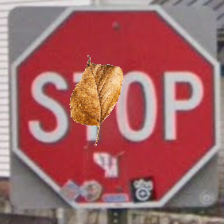}
        \caption{\footnotesize \centering Stop Poplar\newline Attack Image}
    \label{fig:stop_poplar_attack_image}
    \end{subfigure}
    \hfill
    \begin{subfigure}[b]{0.19\textwidth}
        \centering
        \includegraphics[width=2.2cm]{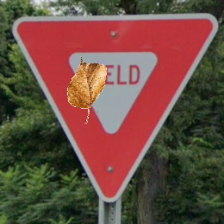}
        \caption{\footnotesize \centering Yield Poplar\newline Attack Image}
    \label{fig:yield_poplar_attack_image}
    \end{subfigure}
    \caption{Adversarial images used in evaluation of the attacks. For each street sign and leaf type, only the adversarial image with highest confidence score is shown, all other images are omitted due to limited space.}
    \label{fig:Attack_Images}
\end{figure*}

Figure~\ref{fig:Test_Images} shows 5 sample street sign images captured under various conditions, with diverse viewing angles and lighting.  Figures~\ref{fig:Output Masks} illustrate our mask generation process. Figure~\ref{fig:Attack_Images} demonstrates the application of the generated leaf masks on these test images, accurately positioning each leaf type to create targeted occlusions on the signs.

\section{Attack Results}
\label{attack_results}

\begin{table}[t]
\centering
\caption{Predicted labels of the adversarial images and their confidence scores.}
\begin{adjustbox}{width=0.48\textwidth}
\label{table_adversarial}
\small
\begin{tabular}{|p{1.8cm}|p{1.8cm}|p{1.8cm}|p{1.8cm}|p{1.2cm}|}
\hline 
\textbf{Adversarial Image} & \textbf{Leaf Type}& \textbf{Predicted Label}  & \textbf{Confidence Score (\%)} & \textbf{Attack Success}
\\ \hline \hline
Yield & Maple & Yield & 70.42 & No \\ \hline
Yield & Oak & Yield & 75.85 & No \\ \hline
Yield & Polar & Yield & 86.96 & No \\ \hline
\hline
Stop & Maple & Ped. Crossing &  59.23 & Yes\\ \hline 
Stop & Oak & Stop & 65.21 & No\\ \hline 
Stop & Polar & Stop & 87.74 & No\\ \hline 
\hline
Turn Right & Maple & Ped. Crossing & 63.45 & Yes\\ \hline 
Turn Right & Oak & Signal Ahead & 42.37 & Yes\\ \hline
Turn Right & Poplar & Added Lane & 37.62 & Yes\\ \hline
\hline
Ped. Crossing  &  Maple & Signal Ahead & 77.90 & Yes\\ \hline 
Ped. Crossing  &  Oak & Signal Ahead & 79.86 & Yes\\ \hline 
Ped. Crossing  &  Poplar & Signal Ahead & 81.71 & Yes\\ \hline 
\hline
Merge & Maple & Ped. Crossing & 97.21 & Yes\\ \hline 
Merge & Oak & Ped. Crossing & 97.14 & Yes\\ \hline 
Merge & Poplar & Ped. Crossing & 96.67 & Yes\\ \hline 
\end{tabular}
\end{adjustbox}
\end{table}

The results of our attack summarized in Table~\ref{table_adversarial}, demonstrate the effectiveness of leaf-based occlusions in inducing misclassifications for certain traffic sign types in the LISA-CNN model. Across different traffic signs, occlusions from maple, oak, and poplar leaves led to successful misclassifications, with confidence scores varying by leaf type and target sign. Notably, for the ``Stop'' sign, only the maple leaf caused a misclassification, predicting ``Pedestrian Crossing'' with a confidence of 59.23\%. 

Similarly, for the ``Turn Right'' sign, all three leaf types induced misclassifications but with different predicted labels and confidence scores, with maple achieving the highest confidence (63.45\%) in predicting ``Pedestrian Crossing.'' The ``Pedestrian Crossing'' and ``Merge'' signs were particularly vulnerable, as each leaf type led to consistent misclassification. For instance, the ``Pedestrian Crossing'' sign was misclassified as ``Signal Ahead'' across all leaf types, with confidence scores ranging from 77.90\% to 81.71\%. The ``Merge'' sign showed the highest vulnerability, with all leaf types leading to a misclassification as ``Pedestrian Crossing'' with confidence scores at or exceeding ~96\%.

However, despite attempts with various leaf placements, both the remaining ``Stop'' images and all ``Yield'' images failed to induce misclassification, highlighting some resilience in certain signs under specific conditions. These results underscore the variability in vulnerability among different traffic signs, depending on leaf type and positioning.

\section{Edge Detection}

Edge detection serves as a fundamental technique for delineating object boundaries in digital images by identifying regions with significant intensity variation. In this work, edge detection is implemented using the Canny algorithm~\cite{rong2014improved}, a multi-step process that combines Gaussian smoothing, gradient computation, and hysteresis thresholding. This method effectively isolates edges within the specified masked region, providing a reliable foundation for further quantitative analysis of edge properties. Table~\ref{table_edge_detection_original} highlights the edge detection metrics used for the test~images and the adversarial images.

The edge {\em length} quantifies the spatial extent of detected edges by summing the pixel lengths of connected edge components within the masked area. This metric is derived through connected components analysis, where each unique edge segment is identified, and its pixel count calculated. The cumulative edge length reflects the continuity and complexity of edge structures, which are essential for analyzing object shape and structure within the region of interest.

\begin{table}[t]
    \centering
    \caption{Edge detection metrics of test images.}
\begin{adjustbox}{width=0.42\textwidth}
\label{table_edge_detection_original}
\small
    \begin{tabular}{|p{1.5cm}|p{1cm}|p{1.5cm}|p{1.5cm}|p{1.5cm}|}
    \hline
        \textbf{Test Image} & \textbf{Edge Length} & \textbf{Orientation} & \textbf{Intensity} & \textbf{Center of Gravity}  \\ \hline
        Stop & 4468 & 2.86 & 142.51 & (133.55, 149.78) \\ \hline
        Pedestrian & 2011 & 0.62 & 101.13 & (89.80, 88.04) \\ \hline
        Yield & 1609 & -1.05 & 134.66 & (131.96, 103.23) \\ \hline
        Right & 1190 & -0.73 & 102.5 & (128.90, 159.93) \\ \hline
         Merge & 889 & 4.89 & 68.8 & (93.94, 96.53) \\ \hline
    \end{tabular}
    \end{adjustbox}
\end{table}

The edge {\em orientation} indicates the predominant directional angle of detected edges, calculated by assessing gradients along the x and y axes via the Sobel operator~\cite{kanopoulos1988design}. By averaging the computed angles of edge pixels, this metric provides an overall orientation of edge features. This information is valuable for characterizing directional patterns, which can support texture classification and directional shape analysis.

The edge {\em intensity} measures the average grayscale intensity of detected edge pixels within the masked area, based on the original image. By calculating the mean brightness of edge pixels, this metric highlights the relative prominence of edges, distinguishing strong edges from weaker transitions. Intensity analysis can be applied in contexts that require emphasis on edge contrast, such as boundary clarity~assessment.

The {\em center of gravity} of detected edges offers a spatial representation of the edge pixel distribution by calculating the mean x and y coordinates of all edge pixels within the masked area. This metric indicates the central tendency of edges, e.g., providing insights into their spatial balance.

\begin{figure*}[t]
\begin{subfigure}[b]{0.19\textwidth}
        \centering
        \includegraphics[width=2.2cm]{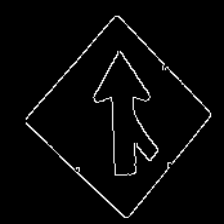}
        \caption{\footnotesize \centering Merge \newline Edge Detection}
    \label{fig:merge_edge_detection}
    \end{subfigure}
    \hfill
\begin{subfigure}[b]{0.19\textwidth}
        \centering
        \includegraphics[width=2.2cm]{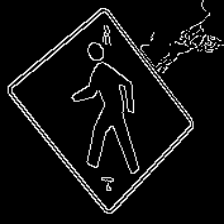}
        \caption{\footnotesize \centering Ped. Crossing \newline Edge Detection}
    \label{fig:ped_crossing_edge_detection}
    \end{subfigure}
    \hfill
\begin{subfigure}[b]{0.19\textwidth}
        \centering
        \includegraphics[width=2.2cm]{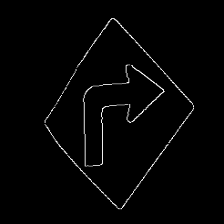}
        \caption{\footnotesize \centering Turn Right \newline Edge Detection}
    \label{fig:right_edge_detection}
    \end{subfigure}
    \hfill
\begin{subfigure}[b]{0.19\textwidth}
        \centering
        \includegraphics[width=2.2cm]{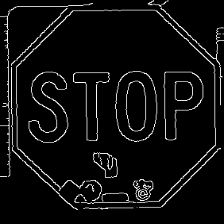}
        \caption{\footnotesize \centering Stop\newline Edge Detection}
    \label{fig:stop_edge_detection}
    \end{subfigure}
    \hfill
\begin{subfigure}[b]{0.19\textwidth}
        \centering
        \includegraphics[width=2.2cm]{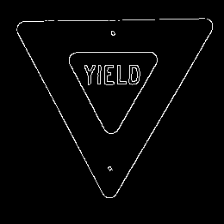}
        \caption{\footnotesize \centering Yield \newline Edge Detection}
    \label{fig:yield_edge_detection}
    \end{subfigure}
    \hfill
\begin{subfigure}[b]{0.19\textwidth}
        \centering
        \includegraphics[width=2.2cm]{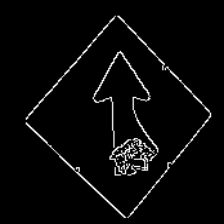}
        \caption{\footnotesize \centering Merge Maple\newline Edge Detection}
    \label{fig:merge_maple_edge_detection}
    \end{subfigure}
    \hfill
 \begin{subfigure}[b]{0.19\textwidth}
        \centering
        \includegraphics[width=2.2cm]{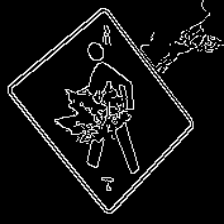}
        \caption{\footnotesize \centering Ped. Crossing Maple\newline Edge Detection}
    \label{fig:ped_crossing_maple_edge_detection}
    \end{subfigure}
    \hfill
\begin{subfigure}[b]{0.19\textwidth}
        \centering
\includegraphics[width=2.2cm]{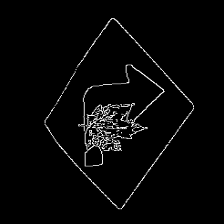}
        \caption{\footnotesize \centering Turn Right Maple \newline Edge Detection}
        \label{fig:right_maple_edge_detection}
    \end{subfigure}
    \hfill
\begin{subfigure}[b]{0.19\textwidth}
        \centering
        \includegraphics[width=2.2cm]{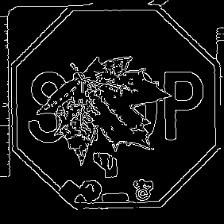}
        \caption{\footnotesize \centering Stop Maple \newline Edge Detection}
    \label{fig:stop_maple_edge_detection}
    \end{subfigure}
    \hfill
\begin{subfigure}[b]{0.19\textwidth}
        \centering
        \includegraphics[width=2.2cm]{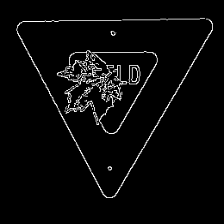}
        \caption{\footnotesize \centering Yield Maple \newline Edge Detection}
    \label{fig:yield_maple_edge_detection}
    \end{subfigure}
    \hfill
 \begin{subfigure}[b]{0.19\textwidth}
        \centering
        \includegraphics[width=2.2cm]{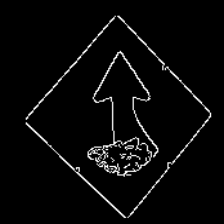}
        \caption{\footnotesize \centering Merge Oak\newline Edge Detection}
    \label{fig:merge_oak_edge_detection}
    \end{subfigure}
    \hfill
\begin{subfigure}[b]{0.19\textwidth}
        \centering
        \includegraphics[width=2.2cm]{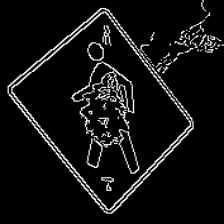}
        \caption{\footnotesize \centering Ped. Crossing Oak\newline Edge Detection}
    \label{fig:ped_crossing_oak_edge_detection}
    \end{subfigure}
    \hfill
 \begin{subfigure}[b]{0.19\textwidth}
        \centering
        \includegraphics[width=2.2cm]{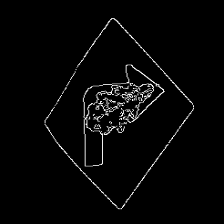}
        \caption{\footnotesize \centering Turn Right Oak\newline Edge Detection}
    \label{fig:right_oak_edge_detection}
    \end{subfigure}
    \hfill
\begin{subfigure}[b]{0.19\textwidth}
        \centering
        \includegraphics[width=2.2cm]{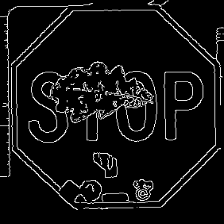}
        \caption{\footnotesize \centering Stop Oak \newline Edge Detection}
    \label{fig:stop_oak_edge_detection}
    \end{subfigure}
    \hfill
\begin{subfigure}[b]{0.19\textwidth}
        \centering
        \includegraphics[width=2.2cm]{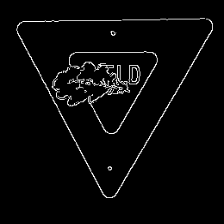}
        \caption{\footnotesize \centering Yield Oak \newline Edge Detection}
    \label{fig:yield_oak_edge_detection}
    \end{subfigure}
    \hfill  
\begin{subfigure}[b]{0.19\textwidth}
        \centering
        \includegraphics[width=2.2cm]{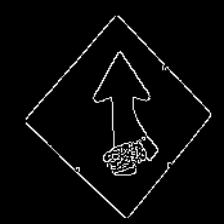}
        \caption{\footnotesize \centering Merge Poplar\newline Edge Detection}
    \label{fig:merge_poplar_edge_detection}
    \end{subfigure}
    \hfill
 \begin{subfigure}[b]{0.19\textwidth}
        \centering
        \includegraphics[width=2.2cm]{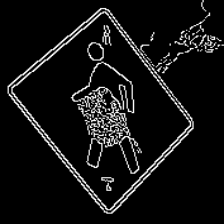}
        \caption{\footnotesize \centering Ped. Crossing Poplar\newline Edge Detection}
\label{fig:ped_crossing_poplar_edge_detection}
    \end{subfigure}
\hfill
 \begin{subfigure}[b]{0.19\textwidth}
        \centering
    \includegraphics[width=2.2cm]{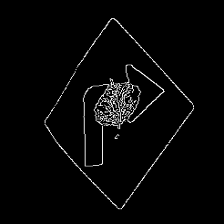}
        \caption{\footnotesize \centering Turn Right Poplar\newline Edge Detection}
    \label{fig:right_poplar_edge_detection}
    \end{subfigure}
    \hfill
 \begin{subfigure}[b]{0.19\textwidth}
        \centering
    \includegraphics[width=2.2cm]{plots/edge_detection_resized/edges_Best_right_poplar_patch_0.4_angle_135.png}
        \caption{\footnotesize \centering Turn Right Poplar\newline Edge Detection}
    \label{fig:right_poplar_edge_detection}
    \end{subfigure}
    \hfill
\begin{subfigure}[b]{0.19\textwidth}
        \centering
        \includegraphics[width=2.2cm]{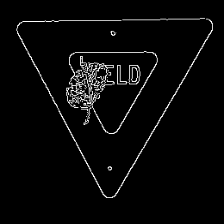}
        \caption{\footnotesize \centering Yield Poplar \newline Edge Detection}
    \label{fig:yield_poplar_edge_detection}
    \end{subfigure}
    \caption{Edge detection of attack images.}
    \label{fig:edge_detection}
\end{figure*}

The values for the edge detection metrics are shown in Table~\ref{table_edge_detection_original}. They can be  compared with the metric values for adversarial images. Edge detection of adversarial images is shown in Figure~\ref{fig:edge_detection}, and the table of metrics for adversarial images is in Table~\ref{table_edge_detection_adversarial}.

\subsection{Analysis of Edge Detection Metrics}

We applied the edge detection to the adversarial images with the aim to understand how different metrics correlate with successful and unsuccessful attacks. Table~\ref{table_edge_detection_adversarial} summarizes the metric values for the adversarial images. At the bottom of the table we list value averages for all successful attacks and all unsuccessful attacks.

To compare the test images to adversarial images, we examine the edge length difference, edge length difference percentage, orientation difference, orientation difference percentage, intensity difference, intensity difference percentage, and the center of gravity distance change. Among these metrics, successful attacks result in higher edge length difference percentage, orientation difference, orientation difference percentage, intensity difference, intensity difference percentage. Meanwhile, unsuccessful attacks have higher edge length difference and longer enter of gravity distance change.

Based on the evaluation, analyzing edge detection metrics could have benefit in detecting adversarial attacks. In particular, both intensity and intensity percentage have higher values for successful attacks. This metric highlights the relative prominence of edges. Intuitively,  when leaves are added, and this has strong impact on change in prominence of edges in the overall image, such images should have higher chance to confuse the classification algorithm, and indeed they do.

\begin{table*}[t]
    \centering
    \caption{Adversarial image edge detection metrics. S denotes a successful attack. U denotes an unsuccessful attack.}
    \label{table_edge_detection_adversarial}
    \begin{adjustbox}{width=\textwidth}
    \small
        \begin{tabular}{|p{1.5cm}|p{1.5cm}|p{1.5cm}|p{1.5cm}|p{1.5cm}|p{1.5cm}|p{1.5cm}|p{1.5cm}|p{1.5cm}|p{1.5cm}|p{1.5cm}|p{1.5cm}|}
            \hline
            \textbf{Adversarial Image} & \textbf{Edge Length} & \textbf{Orientation} & \textbf{Intensity} & \textbf{Center of Gravity} & \textbf{Edge Length Difference} & \textbf{Edge Length Percent} & \textbf{Orientation Difference} & \textbf{Orientation Percent} & \textbf{Intensity Difference} & \textbf{Intensity Percent} & \textbf{Center of Gravity Distance} \\ \hline
             Stop Maple (S) & 6474.00 & 2.69 & 133.17 & (129.65, 147.50) & 2006.00 & 44.90 & 0.17 & 0.05 & 9.34 & 6.55 & 4.52 \\ \hline
             Stop Oak (U) & 5281.00 & 2.41 & 137.43 & (133.53, 146.34) & 813.00 & 18.20 & 0.45 & 0.13 & 5.08 & 3.56 & 3.44 \\ \hline
             Stop Poplar (U) & 5860.00 & 2.56 & 141.32 & (129.93, 144.32) & 1392.00 & 31.15 & 0.30 & 0.08 & 1.19 & 0.84 & 6.55 \\ \hline
            \hline
             Ped. Maple (S) & 2421.00 & 1.28 & 105.48 & (86.98, 90.15) & 410.00 & 20.39 & 0.66 & 0.18 & 4.35 & 4.30 & 3.52 \\ \hline
            Ped. Oak (S) & 2285.00 & 1.62 & 104.04 & (88.98, 88.53) & 274.00 & 13.63 & 1.00 & 0.28 & 2.91 & 2.88 & 0.96 \\ \hline
            Ped. Poplar (S) & 2464.00 & 3.08 & 109.82 & (88.57, 90.89) & 453.00 & 22.53 & 2.46 & 0.68 & 8.69 & 8.59 & 3.10 \\ \hline
            \hline
            Yield Maple (U) & 2452.00 & -3.55 & 124.98 & (122.68, 104.91) & 843.00 & 52.39 & 2.50 & 0.69 & 9.68 & 7.19 & 9.43 \\ \hline
            Yield Oak (U) & 2046.00 & -2.15 & 126.74 & (126.13, 104.92) & 437.00 & 27.16 & 1.10 & 0.31 & 7.92 & 5.88 & 6.07 \\ \hline
            Yield Poplar (U) & 2287.00 & -0.65 & 133.34 & (122.52, 105.32) & 678.00 & 42.14 & 0.40 & 0.11 & 1.32 & 0.98 & 9.67 \\ \hline
            \hline
            Right Maple (S) & 2030.00 & -0.62 & 103.44 & (124.91, 169.36) & 840.00 & 70.59 & 0.11 & 0.03 & 0.94 & 0.92 & 10.24 \\ \hline
             Right Oak (S) & 1960.00 & 1.64 & 102.61 & (129.85, 155.73) & 770.00 & 64.71 & 2.37 & 0.66 & 0.11 & 0.11 & 4.31 \\ \hline
            Right Poplar (S) & 1955.00 & 0.66 & 114.81 & (128.70, 155.64) & 765.00 & 64.29 & 1.39 & 0.39 & 12.31 & 12.01 & 4.29 \\ \hline
            \hline
            Merge Maple (S) & 1071.00 & 3.42 & 75.33 & (95.18, 101.64) & 182.00 & 20.47 & 1.47 & 0.41 & 6.53 & 9.49 & 5.26 \\ \hline
            Merge Oak (S) & 1097.00 & 4.67 & 78.22 & (92.76, 102.36) & 208.00 & 23.40 & 0.22 & 0.06 & 9.42 & 13.69 & 5.95 \\ \hline
            Merge Poplar (S) & 1079.00 & 3.32 & 80.05 & (94.27, 101.95) & 190.00 & 21.37 & 1.57 & 0.44 & 11.25 & 16.35 & 5.43 \\ \hline
            \hline
              & \textbf{Edge Length} & \textbf{Orientation} & \textbf{Intensity} & \textbf{Center of Gravity} & \textbf{Edge Length Difference} & \textbf{Edge Length Percent} & \textbf{Orientation Difference} & \textbf{Orientation Percent} & \textbf{Intensity Difference} & \textbf{Intensity Percent} & \textbf{Center of Gravity Distance} \\ \hline
            \textbf{Average All Successful} & 2283.60 & 2.18 & 100.70 & (105.98, 120.38) & 609.80 & 36.63 & 1.14 & 0.32 & 6.59 & 7.49 & 4.76 \\ \hline
            \textbf{Average All Unsuccessful} & 3585.20 & -0.28 & 132.76 & (126.96, 121.16) & 832.60 & 34.21 & 0.95 & 0.26 & 5.04 & 3.69 & 7.03 \\ \hline
        \end{tabular}
    \end{adjustbox}
\end{table*}

\section{Related Work}
\label{sec:related_work}

This section provides an overview of existing work on existing adversarial example attacks and defenses against them. Our work is first to consider nature-made artifacts, i.e. leaves, for the attack part, while we are first to analyze the correlation between the edge detection and the successful and unsuccessful attacks, which could be used as means for future defenses.

\subsection{Existing Attacks}

The first adversarial attacks on traffic sign recognition (TSR) models emerged in 2017, with notable contributions from Eykholt et al., who developed the Robust Physical Perturbations (RP2) method~\cite{eykholt2018robust}. This approach applies realistic perturbations resembling graffiti on traffic signs, using black-and-white stickers to mislead models by maximizing classification errors. They introduced a two-stage evaluation involving lab tests and real-world "drive-by" scenarios, focusing on attacks against CNN models like LISA-CNN and GTSRB-CNN, which became benchmarks for TSR adversarial studies.

Later, Song et al. extended RP2 for YOLO v2 (You Only Look Once) object detector~\cite{redmon2017yolo9000}, causing detection errors in stop signs by attaching adversarial stickers~\cite{song2018physical}. This method included additional synthetic constraints, such as object rotation, to ensure realistic distortions, but was tested only in stationary lab settings.
Lu et al. critiqued these attacks, showing that advanced detectors like YOLO and Faster-RCNN~\cite{ren2015faster} could withstand them~\cite{lu2017standard}. They eventually introduced modifications allowing adversarial perturbations that could misclassify a stop sign as another object, like a vase. Subsequent studies like Chen et al.’s Shapeshifter~\cite{chen2019shapeshifter} applied Expectation Over Transformation (EOT)~\cite{athalye2018synthesizing} for robustness, targeting stop signs to appear as other objects like people or sports balls.

In black-box attacks, Papernot et al.~\cite{papernot2017practical} first used substitute models trained on observed predictions to generate transferable attacks on TSR models. Woitschek et al.~\cite{woitschek2021physical} later examined black-box techniques, including gradient approximation, while Li et al.~\cite{li2020adaptive} introduced the adaptive square attack, which updates perturbations in random squares to bypass model defenses.

Alternative methods include ``innocuous-looking'' attacks, where signs like advertisements are modified to appear as traffic signs. Sitawarin et al.~\cite{sitawarin2018rogue} used Carlini
and Wagner (C\&W)~\cite{carlini2017towards} attacks combined with EOT to generate such adversarial signs, leading to subsequent developments like DARTS~\cite{sitawarin2018darts}, which introduced lenticular printing techniques for attacks viewed differently from various angles.

Recent approaches, such as those by Liu et al., utilize generative adversarial networks
(GANs) and attention mechanisms to create naturalistic adversarial stickers~\cite{liu2019perceptual}. Methods imitating raindrops~\cite{liu2023adversarial},  shadows~\cite{zhong2022shadows} or light~\cite{hsiao2024natural} further enhance stealth, targeting visually susceptible areas on traffic signs and disrupt the model's performance. These techniques exploit vulnerabilities in vision models, allowing subtle modifications that significantly impact traffic sign classification.

\subsection{Existing Defenses}

Knowledge distillation, introduced by Hinton et al.\cite{hinton2015distilling}, provides a technique to transfer insights from more complex neural networks to simpler ones. Extending this concept, Papernot et al.\cite{papernot2016distillation} developed defensive distillation as a countermeasure to adversarial attacks. This method initially trains a distillation model using standard inputs and labels to generate a soft probability distribution. These probabilities, combined with the original examples, are used to train a secondary model of identical architecture, producing an updated probability distribution as labels that fortify the model’s resistance to adversarial interference.

To thwart adversaries from leveraging gradient information, Folz et al.~\cite{folz2020adversarial} proposed S2SNet, a defense mechanism that obfuscates the model's gradient. S2SNet encodes category-relevant information into structural data that impacts the gradient, preserving only the core structural elements essential for classification while discarding irrelevant details to reduce vulnerability to adversarial perturbations. 
Liao et al.~\cite{liao2018defense} presented the high-level representation guided denoiser (HGD), employing a U-Net architecture as the denoising component. This strategy incorporates a feature-focused loss function that mitigates error amplification, thereby enhancing model robustness and performance under adversarial scenarios.

Wu et al.\cite{wu2019defending} explored the limitations of adversarial training with PGD attacks\cite{madry2017towards} and randomized smoothing when facing sophisticated physical attacks. They introduced Rectangular Occlusion Attacks (ROA), which involve embedding a small adversarial rectangle into images, and showed that adversarial training against ROA substantially strengthens image classifiers against physical attacks.

Finally, Cohen et al.~\cite{cohen2020detecting} developed a detection approach for adversarial attacks on pre-trained neural networks. By applying influence functions and k-nearest neighbor (k-NN) analysis on activation layers, they identified supportive training samples. Regular inputs demonstrated high similarity to nearest neighbors, while adversarial inputs diverged, enabling detection of adversarial examples.

\section{Conclusion}
\label{sec_conclusion}

This work introduced a novel class of adversarial attacks in image classification systems by demonstrating how nature-made artifacts, specifically tree leaves, can be exploited to cause misclassification in street sign recognition—a critical component in autonomous automotive systems. Unlike traditional adversarial attacks that rely on human-made modifications, such as stickers or paint, this approach leverages the natural plausibility of fall leaves adhering to signs, providing attackers with a layer of deniability and making detection and mitigation more challenging. Through comprehensive evaluation across various leaf types, sizes, colors, and orientations, the study reveals a high success rate of misclassification, underscoring the vulnerability of traffic sign recognition models to such attacks. To help understand the attacks, the work included the analysis of correlation between successful misclassification and changes in different edge detection metrics, and demonstrates different correlation of the metrics to successful and unsuccessful attacks. This study ultimately calls for future research into defense mechanisms to mitigate the impact of nature-based adversarial perturbations on critical image classification tasks.

\section*{Acknowledgements}

This work was supported in part by National Science Foundation grant \nsf{2245344}.

\bibliographystyle{ACM-Reference-Format}
\bibliography{bibtex/references}


\begin{thebibliography}{40}


\ifx \showCODEN    \undefined \def \showCODEN     #1{\unskip}     \fi
\ifx \showDOI      \undefined \def \showDOI       #1{#1}\fi
\ifx \showISBNx    \undefined \def \showISBNx     #1{\unskip}     \fi
\ifx \showISBNxiii \undefined \def \showISBNxiii  #1{\unskip}     \fi
\ifx \showISSN     \undefined \def \showISSN      #1{\unskip}     \fi
\ifx \showLCCN     \undefined \def \showLCCN      #1{\unskip}     \fi
\ifx \shownote     \undefined \def \shownote      #1{#1}          \fi
\ifx \showarticletitle \undefined \def \showarticletitle #1{#1}   \fi
\ifx \showURL      \undefined \def \showURL       {\relax}        \fi
\providecommand\bibfield[2]{#2}
\providecommand\bibinfo[2]{#2}
\providecommand\natexlab[1]{#1}
\providecommand\showeprint[2][]{arXiv:#2}

\bibitem[Athalye et~al\mbox{.}(2018)]%
        {athalye2018synthesizing}
\bibfield{author}{\bibinfo{person}{Anish Athalye}, \bibinfo{person}{Logan Engstrom}, \bibinfo{person}{Andrew Ilyas}, {and} \bibinfo{person}{Kevin Kwok}.} \bibinfo{year}{2018}\natexlab{}.
\newblock \showarticletitle{Synthesizing robust adversarial examples}. In \bibinfo{booktitle}{\emph{International conference on machine learning}}. PMLR, \bibinfo{pages}{284--293}.
\newblock


\bibitem[Brown et~al\mbox{.}(2017)]%
        {brown2017adversarial}
\bibfield{author}{\bibinfo{person}{Tom~B Brown}, \bibinfo{person}{Dandelion Man{\'e}}, \bibinfo{person}{Aurko Roy}, \bibinfo{person}{Mart{\'\i}n Abadi}, {and} \bibinfo{person}{Justin Gilmer}.} \bibinfo{year}{2017}\natexlab{}.
\newblock \showarticletitle{Adversarial patch}.
\newblock \bibinfo{journal}{\emph{arXiv preprint arXiv:1712.09665}} (\bibinfo{year}{2017}).
\newblock


\bibitem[Carlini and Wagner(2017)]%
        {carlini2017towards}
\bibfield{author}{\bibinfo{person}{Nicholas Carlini} {and} \bibinfo{person}{David Wagner}.} \bibinfo{year}{2017}\natexlab{}.
\newblock \showarticletitle{Towards evaluating the robustness of neural networks}. In \bibinfo{booktitle}{\emph{2017 ieee symposium on security and privacy (sp)}}. Ieee, \bibinfo{pages}{39--57}.
\newblock


\bibitem[Chakraborty et~al\mbox{.}(2018)]%
        {chakraborty2018adversarial}
\bibfield{author}{\bibinfo{person}{Anirban Chakraborty}, \bibinfo{person}{Manaar Alam}, \bibinfo{person}{Vishal Dey}, \bibinfo{person}{Anupam Chattopadhyay}, {and} \bibinfo{person}{Debdeep Mukhopadhyay}.} \bibinfo{year}{2018}\natexlab{}.
\newblock \showarticletitle{Adversarial attacks and defences: A survey}.
\newblock \bibinfo{journal}{\emph{arXiv preprint arXiv:1810.00069}} (\bibinfo{year}{2018}).
\newblock


\bibitem[Chen et~al\mbox{.}(2019)]%
        {chen2019shapeshifter}
\bibfield{author}{\bibinfo{person}{Shang-Tse Chen}, \bibinfo{person}{Cory Cornelius}, \bibinfo{person}{Jason Martin}, {and} \bibinfo{person}{Duen~Horng Chau}.} \bibinfo{year}{2019}\natexlab{}.
\newblock \showarticletitle{Shapeshifter: Robust physical adversarial attack on faster r-cnn object detector}. In \bibinfo{booktitle}{\emph{Machine Learning and Knowledge Discovery in Databases: European Conference, ECML PKDD 2018, Dublin, Ireland, September 10--14, 2018, Proceedings, Part I 18}}. Springer, \bibinfo{pages}{52--68}.
\newblock


\bibitem[Cohen et~al\mbox{.}(2020)]%
        {cohen2020detecting}
\bibfield{author}{\bibinfo{person}{Gilad Cohen}, \bibinfo{person}{Guillermo Sapiro}, {and} \bibinfo{person}{Raja Giryes}.} \bibinfo{year}{2020}\natexlab{}.
\newblock \showarticletitle{Detecting adversarial samples using influence functions and nearest neighbors}. In \bibinfo{booktitle}{\emph{Proceedings of the IEEE/CVF conference on computer vision and pattern recognition}}. \bibinfo{pages}{14453--14462}.
\newblock


\bibitem[Eykholt et~al\mbox{.}(2018)]%
        {eykholt2018robust}
\bibfield{author}{\bibinfo{person}{Kevin Eykholt}, \bibinfo{person}{Ivan Evtimov}, \bibinfo{person}{Earlence Fernandes}, \bibinfo{person}{Bo Li}, \bibinfo{person}{Amir Rahmati}, \bibinfo{person}{Chaowei Xiao}, \bibinfo{person}{Atul Prakash}, \bibinfo{person}{Tadayoshi Kohno}, {and} \bibinfo{person}{Dawn Song}.} \bibinfo{year}{2018}\natexlab{}.
\newblock \showarticletitle{Robust physical-world attacks on deep learning visual classification}. In \bibinfo{booktitle}{\emph{Proceedings of the IEEE conference on computer vision and pattern recognition}}. \bibinfo{pages}{1625--1634}.
\newblock


\bibitem[Folz et~al\mbox{.}(2020)]%
        {folz2020adversarial}
\bibfield{author}{\bibinfo{person}{Joachim Folz}, \bibinfo{person}{Sebastian Palacio}, \bibinfo{person}{Joern Hees}, {and} \bibinfo{person}{Andreas Dengel}.} \bibinfo{year}{2020}\natexlab{}.
\newblock \showarticletitle{Adversarial defense based on structure-to-signal autoencoders}. In \bibinfo{booktitle}{\emph{2020 IEEE Winter Conference on Applications of Computer Vision (WACV)}}. IEEE, \bibinfo{pages}{3568--3577}.
\newblock


\bibitem[Gil and Kimmel(2002)]%
        {gil2002efficient}
\bibfield{author}{\bibinfo{person}{Joseph~Yossi Gil} {and} \bibinfo{person}{Ron Kimmel}.} \bibinfo{year}{2002}\natexlab{}.
\newblock \showarticletitle{Efficient dilation, erosion, opening, and closing algorithms}.
\newblock \bibinfo{journal}{\emph{IEEE Transactions on Pattern Analysis and Machine Intelligence}} \bibinfo{volume}{24}, \bibinfo{number}{12} (\bibinfo{year}{2002}), \bibinfo{pages}{1606--1617}.
\newblock


\bibitem[Goodfellow et~al\mbox{.}(2014)]%
        {goodfellow2014explaining}
\bibfield{author}{\bibinfo{person}{Ian~J Goodfellow}, \bibinfo{person}{Jonathon Shlens}, {and} \bibinfo{person}{Christian Szegedy}.} \bibinfo{year}{2014}\natexlab{}.
\newblock \showarticletitle{Explaining and harnessing adversarial examples}.
\newblock \bibinfo{journal}{\emph{arXiv preprint arXiv:1412.6572}} (\bibinfo{year}{2014}).
\newblock


\bibitem[Hinton(2015)]%
        {hinton2015distilling}
\bibfield{author}{\bibinfo{person}{Geoffrey Hinton}.} \bibinfo{year}{2015}\natexlab{}.
\newblock \showarticletitle{Distilling the Knowledge in a Neural Network}.
\newblock \bibinfo{journal}{\emph{arXiv preprint arXiv:1503.02531}} (\bibinfo{year}{2015}).
\newblock


\bibitem[Hsiao et~al\mbox{.}(2024)]%
        {hsiao2024natural}
\bibfield{author}{\bibinfo{person}{Teng-Fang Hsiao}, \bibinfo{person}{Bo-Lun Huang}, \bibinfo{person}{Zi-Xiang Ni}, \bibinfo{person}{Yan-Ting Lin}, \bibinfo{person}{Hong-Han Shuai}, \bibinfo{person}{Yung-Hui Li}, {and} \bibinfo{person}{Wen-Huang Cheng}.} \bibinfo{year}{2024}\natexlab{}.
\newblock \showarticletitle{Natural Light Can Also be Dangerous: Traffic Sign Misinterpretation Under Adversarial Natural Light Attacks}. In \bibinfo{booktitle}{\emph{Proceedings of the IEEE/CVF Winter Conference on Applications of Computer Vision}}. \bibinfo{pages}{3915--3924}.
\newblock


\bibitem[Hummel et~al\mbox{.}(1987)]%
        {hummel1987deblurring}
\bibfield{author}{\bibinfo{person}{Robert~A Hummel}, \bibinfo{person}{B Kimia}, {and} \bibinfo{person}{Steven~W Zucker}.} \bibinfo{year}{1987}\natexlab{}.
\newblock \showarticletitle{Deblurring gaussian blur}.
\newblock \bibinfo{journal}{\emph{Computer Vision, Graphics, and Image Processing}} \bibinfo{volume}{38}, \bibinfo{number}{1} (\bibinfo{year}{1987}), \bibinfo{pages}{66--80}.
\newblock


\bibitem[Kanopoulos et~al\mbox{.}(1988)]%
        {kanopoulos1988design}
\bibfield{author}{\bibinfo{person}{Nick Kanopoulos}, \bibinfo{person}{Nagesh Vasanthavada}, {and} \bibinfo{person}{Robert~L Baker}.} \bibinfo{year}{1988}\natexlab{}.
\newblock \showarticletitle{Design of an image edge detection filter using the Sobel operator}.
\newblock \bibinfo{journal}{\emph{IEEE Journal of solid-state circuits}} \bibinfo{volume}{23}, \bibinfo{number}{2} (\bibinfo{year}{1988}), \bibinfo{pages}{358--367}.
\newblock


\bibitem[Li et~al\mbox{.}(2020)]%
        {li2020adaptive}
\bibfield{author}{\bibinfo{person}{Yujie Li}, \bibinfo{person}{Xing Xu}, \bibinfo{person}{Jinhui Xiao}, \bibinfo{person}{Siyuan Li}, {and} \bibinfo{person}{Heng~Tao Shen}.} \bibinfo{year}{2020}\natexlab{}.
\newblock \showarticletitle{Adaptive square attack: Fooling autonomous cars with adversarial traffic signs}.
\newblock \bibinfo{journal}{\emph{IEEE Internet of Things Journal}} \bibinfo{volume}{8}, \bibinfo{number}{8} (\bibinfo{year}{2020}), \bibinfo{pages}{6337--6347}.
\newblock


\bibitem[Liang et~al\mbox{.}(2022)]%
        {liang2022adversarial}
\bibfield{author}{\bibinfo{person}{Hongshuo Liang}, \bibinfo{person}{Erlu He}, \bibinfo{person}{Yangyang Zhao}, \bibinfo{person}{Zhe Jia}, {and} \bibinfo{person}{Hao Li}.} \bibinfo{year}{2022}\natexlab{}.
\newblock \showarticletitle{Adversarial attack and defense: A survey}.
\newblock \bibinfo{journal}{\emph{Electronics}} \bibinfo{volume}{11}, \bibinfo{number}{8} (\bibinfo{year}{2022}), \bibinfo{pages}{1283}.
\newblock


\bibitem[Liao et~al\mbox{.}(2018)]%
        {liao2018defense}
\bibfield{author}{\bibinfo{person}{Fangzhou Liao}, \bibinfo{person}{Ming Liang}, \bibinfo{person}{Yinpeng Dong}, \bibinfo{person}{Tianyu Pang}, \bibinfo{person}{Xiaolin Hu}, {and} \bibinfo{person}{Jun Zhu}.} \bibinfo{year}{2018}\natexlab{}.
\newblock \showarticletitle{Defense against adversarial attacks using high-level representation guided denoiser}. In \bibinfo{booktitle}{\emph{Proceedings of the IEEE conference on computer vision and pattern recognition}}. \bibinfo{pages}{1778--1787}.
\newblock


\bibitem[Liashchynskyi and Liashchynskyi(2019)]%
        {liashchynskyi2019grid}
\bibfield{author}{\bibinfo{person}{Petro Liashchynskyi} {and} \bibinfo{person}{Pavlo Liashchynskyi}.} \bibinfo{year}{2019}\natexlab{}.
\newblock \showarticletitle{Grid search, random search, genetic algorithm: a big comparison for NAS}.
\newblock \bibinfo{journal}{\emph{arXiv preprint arXiv:1912.06059}} (\bibinfo{year}{2019}).
\newblock


\bibitem[Liu et~al\mbox{.}(2019)]%
        {liu2019perceptual}
\bibfield{author}{\bibinfo{person}{Aishan Liu}, \bibinfo{person}{Xianglong Liu}, \bibinfo{person}{Jiaxin Fan}, \bibinfo{person}{Yuqing Ma}, \bibinfo{person}{Anlan Zhang}, \bibinfo{person}{Huiyuan Xie}, {and} \bibinfo{person}{Dacheng Tao}.} \bibinfo{year}{2019}\natexlab{}.
\newblock \showarticletitle{Perceptual-sensitive gan for generating adversarial patches}. In \bibinfo{booktitle}{\emph{Proceedings of the AAAI conference on artificial intelligence}}, Vol.~\bibinfo{volume}{33}. \bibinfo{pages}{1028--1035}.
\newblock


\bibitem[Liu et~al\mbox{.}(2023)]%
        {liu2023adversarial}
\bibfield{author}{\bibinfo{person}{Jiyuan Liu}, \bibinfo{person}{Bingyi Lu}, \bibinfo{person}{Mingkang Xiong}, \bibinfo{person}{Tao Zhang}, {and} \bibinfo{person}{Huilin Xiong}.} \bibinfo{year}{2023}\natexlab{}.
\newblock \showarticletitle{Adversarial attack with raindrops}.
\newblock \bibinfo{journal}{\emph{arXiv preprint arXiv:2302.14267}} (\bibinfo{year}{2023}).
\newblock


\bibitem[Lu et~al\mbox{.}(2017a)]%
        {lu2017adversarial}
\bibfield{author}{\bibinfo{person}{Jiajun Lu}, \bibinfo{person}{Hussein Sibai}, {and} \bibinfo{person}{Evan Fabry}.} \bibinfo{year}{2017}\natexlab{a}.
\newblock \showarticletitle{Adversarial examples that fool detectors}.
\newblock \bibinfo{journal}{\emph{arXiv preprint arXiv:1712.02494}} (\bibinfo{year}{2017}).
\newblock


\bibitem[Lu et~al\mbox{.}(2017b)]%
        {lu2017standard}
\bibfield{author}{\bibinfo{person}{Jiajun Lu}, \bibinfo{person}{Hussein Sibai}, \bibinfo{person}{Evan Fabry}, {and} \bibinfo{person}{David Forsyth}.} \bibinfo{year}{2017}\natexlab{b}.
\newblock \showarticletitle{Standard detectors aren't (currently) fooled by physical adversarial stop signs}.
\newblock \bibinfo{journal}{\emph{arXiv preprint arXiv:1710.03337}} (\bibinfo{year}{2017}).
\newblock


\bibitem[Madry(2017)]%
        {madry2017towards}
\bibfield{author}{\bibinfo{person}{Aleksander Madry}.} \bibinfo{year}{2017}\natexlab{}.
\newblock \showarticletitle{Towards deep learning models resistant to adversarial attacks}.
\newblock \bibinfo{journal}{\emph{arXiv preprint arXiv:1706.06083}} (\bibinfo{year}{2017}).
\newblock


\bibitem[M{\k{a}}dry et~al\mbox{.}(2017)]%
        {mkadry2017towards}
\bibfield{author}{\bibinfo{person}{Aleksander M{\k{a}}dry}, \bibinfo{person}{Aleksandar Makelov}, \bibinfo{person}{Ludwig Schmidt}, \bibinfo{person}{Dimitris Tsipras}, {and} \bibinfo{person}{Adrian Vladu}.} \bibinfo{year}{2017}\natexlab{}.
\newblock \showarticletitle{Towards deep learning models resistant to adversarial attacks}.
\newblock \bibinfo{journal}{\emph{stat}} \bibinfo{volume}{1050}, \bibinfo{number}{9} (\bibinfo{year}{2017}).
\newblock


\bibitem[Mogelmose et~al\mbox{.}(2012)]%
        {lisa}
\bibfield{author}{\bibinfo{person}{Andreas Mogelmose}, \bibinfo{person}{Mohan~Manubhai Trivedi}, {and} \bibinfo{person}{Thomas~B Moeslund}.} \bibinfo{year}{2012}\natexlab{}.
\newblock \showarticletitle{Vision-based traffic sign detection and analysis for intelligent driver assistance systems: Perspectives and survey}.
\newblock \bibinfo{journal}{\emph{IEEE transactions on intelligent transportation systems}} \bibinfo{volume}{13}, \bibinfo{number}{4} (\bibinfo{year}{2012}), \bibinfo{pages}{1484--1497}.
\newblock


\bibitem[Moosavi-Dezfooli et~al\mbox{.}(2017)]%
        {moosavi2017universal}
\bibfield{author}{\bibinfo{person}{Seyed-Mohsen Moosavi-Dezfooli}, \bibinfo{person}{Alhussein Fawzi}, \bibinfo{person}{Omar Fawzi}, {and} \bibinfo{person}{Pascal Frossard}.} \bibinfo{year}{2017}\natexlab{}.
\newblock \showarticletitle{Universal adversarial perturbations}. In \bibinfo{booktitle}{\emph{Proceedings of the IEEE conference on computer vision and pattern recognition}}. \bibinfo{pages}{1765--1773}.
\newblock


\bibitem[Papernot et~al\mbox{.}(2017)]%
        {papernot2017practical}
\bibfield{author}{\bibinfo{person}{Nicolas Papernot}, \bibinfo{person}{Patrick McDaniel}, \bibinfo{person}{Ian Goodfellow}, \bibinfo{person}{Somesh Jha}, \bibinfo{person}{Z~Berkay Celik}, {and} \bibinfo{person}{Ananthram Swami}.} \bibinfo{year}{2017}\natexlab{}.
\newblock \showarticletitle{Practical black-box attacks against machine learning}. In \bibinfo{booktitle}{\emph{Proceedings of the 2017 ACM on Asia conference on computer and communications security}}. \bibinfo{pages}{506--519}.
\newblock


\bibitem[Papernot et~al\mbox{.}(2016)]%
        {papernot2016distillation}
\bibfield{author}{\bibinfo{person}{Nicolas Papernot}, \bibinfo{person}{Patrick McDaniel}, \bibinfo{person}{Xi Wu}, \bibinfo{person}{Somesh Jha}, {and} \bibinfo{person}{Ananthram Swami}.} \bibinfo{year}{2016}\natexlab{}.
\newblock \showarticletitle{Distillation as a defense to adversarial perturbations against deep neural networks}. In \bibinfo{booktitle}{\emph{2016 IEEE symposium on security and privacy (SP)}}. IEEE, \bibinfo{pages}{582--597}.
\newblock


\bibitem[Pavlitska et~al\mbox{.}(2023)]%
        {pavlitska2023adversarial}
\bibfield{author}{\bibinfo{person}{Svetlana Pavlitska}, \bibinfo{person}{Nico Lambing}, {and} \bibinfo{person}{J~Marius Z{\"o}llner}.} \bibinfo{year}{2023}\natexlab{}.
\newblock \showarticletitle{Adversarial attacks on traffic sign recognition: A survey}. In \bibinfo{booktitle}{\emph{International Conference on Electrical, Computer, Communications and Mechatronics Engineering}}. IEEE, \bibinfo{pages}{1--6}.
\newblock


\bibitem[Redmon and Farhadi(2017)]%
        {redmon2017yolo9000}
\bibfield{author}{\bibinfo{person}{Joseph Redmon} {and} \bibinfo{person}{Ali Farhadi}.} \bibinfo{year}{2017}\natexlab{}.
\newblock \showarticletitle{YOLO9000: better, faster, stronger}. In \bibinfo{booktitle}{\emph{Proceedings of the IEEE conference on computer vision and pattern recognition}}. \bibinfo{pages}{7263--7271}.
\newblock


\bibitem[Ren(2015)]%
        {ren2015faster}
\bibfield{author}{\bibinfo{person}{Shaoqing Ren}.} \bibinfo{year}{2015}\natexlab{}.
\newblock \showarticletitle{Faster r-cnn: Towards real-time object detection with region proposal networks}.
\newblock \bibinfo{journal}{\emph{arXiv preprint arXiv:1506.01497}} (\bibinfo{year}{2015}).
\newblock


\bibitem[Rong et~al\mbox{.}(2014)]%
        {rong2014improved}
\bibfield{author}{\bibinfo{person}{Weibin Rong}, \bibinfo{person}{Zhanjing Li}, \bibinfo{person}{Wei Zhang}, {and} \bibinfo{person}{Lining Sun}.} \bibinfo{year}{2014}\natexlab{}.
\newblock \showarticletitle{An improved CANNY edge detection algorithm}. In \bibinfo{booktitle}{\emph{2014 IEEE international conference on mechatronics and automation}}. IEEE, \bibinfo{pages}{577--582}.
\newblock


\bibitem[Sitawarin et~al\mbox{.}(2018a)]%
        {sitawarin2018darts}
\bibfield{author}{\bibinfo{person}{Chawin Sitawarin}, \bibinfo{person}{Arjun~Nitin Bhagoji}, \bibinfo{person}{Arsalan Mosenia}, \bibinfo{person}{Mung Chiang}, {and} \bibinfo{person}{Prateek Mittal}.} \bibinfo{year}{2018}\natexlab{a}.
\newblock \showarticletitle{Darts: Deceiving autonomous cars with toxic signs}.
\newblock \bibinfo{journal}{\emph{arXiv preprint arXiv:1802.06430}} (\bibinfo{year}{2018}).
\newblock


\bibitem[Sitawarin et~al\mbox{.}(2018b)]%
        {sitawarin2018rogue}
\bibfield{author}{\bibinfo{person}{Chawin Sitawarin}, \bibinfo{person}{Arjun~Nitin Bhagoji}, \bibinfo{person}{Arsalan Mosenia}, \bibinfo{person}{Prateek Mittal}, {and} \bibinfo{person}{Mung Chiang}.} \bibinfo{year}{2018}\natexlab{b}.
\newblock \showarticletitle{Rogue signs: Deceiving traffic sign recognition with malicious ads and logos}.
\newblock \bibinfo{journal}{\emph{arXiv preprint arXiv:1801.02780}} (\bibinfo{year}{2018}).
\newblock


\bibitem[Song et~al\mbox{.}(2018)]%
        {song2018physical}
\bibfield{author}{\bibinfo{person}{Dawn Song}, \bibinfo{person}{Kevin Eykholt}, \bibinfo{person}{Ivan Evtimov}, \bibinfo{person}{Earlence Fernandes}, \bibinfo{person}{Bo Li}, \bibinfo{person}{Amir Rahmati}, \bibinfo{person}{Florian Tramer}, \bibinfo{person}{Atul Prakash}, {and} \bibinfo{person}{Tadayoshi Kohno}.} \bibinfo{year}{2018}\natexlab{}.
\newblock \showarticletitle{Physical adversarial examples for object detectors}. In \bibinfo{booktitle}{\emph{12th USENIX workshop on offensive technologies (WOOT 18)}}.
\newblock


\bibitem[Szegedy(2013)]%
        {szegedy2013intriguing}
\bibfield{author}{\bibinfo{person}{C Szegedy}.} \bibinfo{year}{2013}\natexlab{}.
\newblock \showarticletitle{Intriguing properties of neural networks}.
\newblock \bibinfo{journal}{\emph{arXiv preprint arXiv:1312.6199}} (\bibinfo{year}{2013}).
\newblock


\bibitem[Wei et~al\mbox{.}(2022)]%
        {wei2022adversarial}
\bibfield{author}{\bibinfo{person}{Xingxing Wei}, \bibinfo{person}{Ying Guo}, {and} \bibinfo{person}{Jie Yu}.} \bibinfo{year}{2022}\natexlab{}.
\newblock \showarticletitle{Adversarial sticker: A stealthy attack method in the physical world}.
\newblock \bibinfo{journal}{\emph{IEEE Transactions on Pattern Analysis and Machine Intelligence}} \bibinfo{volume}{45}, \bibinfo{number}{3} (\bibinfo{year}{2022}), \bibinfo{pages}{2711--2725}.
\newblock


\bibitem[Woitschek and Schneider(2021)]%
        {woitschek2021physical}
\bibfield{author}{\bibinfo{person}{Fabian Woitschek} {and} \bibinfo{person}{Georg Schneider}.} \bibinfo{year}{2021}\natexlab{}.
\newblock \showarticletitle{Physical adversarial attacks on deep neural networks for traffic sign recognition: A feasibility study}. In \bibinfo{booktitle}{\emph{2021 IEEE Intelligent vehicles symposium (IV)}}. IEEE, \bibinfo{pages}{481--487}.
\newblock


\bibitem[Wu et~al\mbox{.}(2019)]%
        {wu2019defending}
\bibfield{author}{\bibinfo{person}{Tong Wu}, \bibinfo{person}{Liang Tong}, {and} \bibinfo{person}{Yevgeniy Vorobeychik}.} \bibinfo{year}{2019}\natexlab{}.
\newblock \showarticletitle{Defending against physically realizable attacks on image classification}.
\newblock \bibinfo{journal}{\emph{arXiv preprint arXiv:1909.09552}} (\bibinfo{year}{2019}).
\newblock


\bibitem[Zhong et~al\mbox{.}(2022)]%
        {zhong2022shadows}
\bibfield{author}{\bibinfo{person}{Yiqi Zhong}, \bibinfo{person}{Xianming Liu}, \bibinfo{person}{Deming Zhai}, \bibinfo{person}{Junjun Jiang}, {and} \bibinfo{person}{Xiangyang Ji}.} \bibinfo{year}{2022}\natexlab{}.
\newblock \showarticletitle{Shadows can be dangerous: Stealthy and effective physical-world adversarial attack by natural phenomenon}. In \bibinfo{booktitle}{\emph{Proceedings of the IEEE/CVF Conference on Computer Vision and Pattern Recognition}}. \bibinfo{pages}{15345--15354}.
\newblock


\end{thebibliography}

\end{document}